\documentclass[runningheads]{llncs}
\usepackage[T1]{fontenc}
\usepackage[
  left=3cm,
  right=3cm,
  top=3cm,
  bottom=3cm
]{geometry}
\usepackage{graphicx}
\usepackage{booktabs}
\usepackage{amsmath}
\usepackage{amssymb}
\usepackage{mathrsfs}
\usepackage{adjustbox}
\usepackage{tikz}
\usetikzlibrary{arrows.meta, positioning}
\usepackage{hyperref}

\usepackage{bbm} 
\usepackage{stmaryrd} 

\usepackage{multirow}
\usepackage{makecell}
\usepackage{adjustbox} 

\usepackage{xcolor} 
\usepackage{comment} 
\usepackage[misc]{ifsym}
\newcommand{\corr}{(\Letter)}
\usepackage{float}

\begin{document}

\title{MultiSigBERT: Beyond Survival Analysis through Multimodal and Sequential Modeling in Oncology}

\titlerunning{MultiSigBERT: Multimodal and Sequential Survival Modeling}


\author{Paul Minchella\inst{1} \corr  \and Stéphane Chrétien \inst{1} \and Guillaume Metzler \inst{1} \and Loic Verlingue\inst{2} \and Rémi Vaucher \inst{3}}



\authorrunning{P. Minchella et al.}


\institute{Université Lumière Lyon 2, 69500 Bron, France
\email{\{paul.minchella, stephane.chretien, guillaume.metzler\}@univ-lyon2.fr}
\and
Léon Bérard Center, Lyon 69008, France \email{loic.verlingue@lyon.unicancer.fr}
\and EPITA, Lyon 69007, France
\email{remi.vaucher@epita.fr}
}

\toctitle{MultiSigBERT: Beyond Survival Analysis through Multimodal and Sequential Modeling in Oncology}
\tocauthor{Paul Minchella, Loic Verlingue, Stéphane Chrétien, Rémi Vaucher, Guillaume Metzler}

\maketitle              

\begin{abstract}
Machine learning has become an essential component of modern healthcare, where the integration of heterogeneous data sources offers unprecedented opportunities to improve clinical decision-making. Electronic Health Records (EHR) contain complementary information -- including narrative clinical reports, numerical measurements, and structured variables -- yet most survival models remain limited to a single modality or fail to exploit the temporal nature of patient trajectories.
We propose MultiSigBERT, a unified framework for multimodal sequential survival modeling in oncology based on path signature representations. Here, narrative medical reports (free-text) are converted into sentence embeddings by extracting and averaging contextual word embeddings. These representations are then compressed via modality-specific PCA and concatenated with structured covariates to form joint temporal trajectories which are then encoded using the Signature transform, a tool from Rough Paths theory that efficiently captures higher-order temporal interactions across modalities without supervision needed. 
The  computed Signature features are finally  incorporated as high dimensional features into a LASSO-regularized Cox model to estimate individualized risk scores.

The performance of our novel MultiSigBERT pipeline is illustrated on the analysis of a real-world oncology cohort from the Léon Bérard Center, comprising over 120,000 medical reports and structured records from more than 2,500 patients. The model achieves a concordance index of 0.743 (sd 0.029). on an independent test set, demonstrating the benefit of jointly modeling multimodal temporal dynamics together with patient-level geometric structure for survival prediction.

\end{abstract}

\section{Introduction}
\subsection{Background and Challenges}

Survival analysis aims to model the time until an event of interest occurs, such as death, recurrence, or disease progression. In oncology, accurate survival prediction is essential for treatment selection, clinical trial inclusion, and follow-up planning.

Modern electronic health records (EHRs) contain heterogeneous and complementary information, including structured clinical variables, laboratory measurements, molecular data, and narrative medical reports. This diversity offers strong potential to improve survival modeling. However, integrating these data sources remains challenging. Most classical survival models rely on static baseline covariates and ignore the longitudinal nature of patient follow-up. Even recent machine learning approaches often struggle with irregular, asynchronous, and time-dependent clinical data.

In addition, many existing methods are unimodal, focusing either on structured variables or unstructured text. Such approaches fail to capture the interactions between modalities that shape patient trajectories and influence prognosis.

To address these limitations, we introduce MultiSigBERT, a multimodal and sequential survival framework designed to integrate heterogeneous clinical data while modeling temporal dynamics in a principled and interpretable manner. By combining embedding representations, rough path signatures, graph-based clustering, and sparse survival modeling, this model provides a unified approach for analyzing real-world multimodal oncology data.

\subsection{Related Works}

Survival analysis is a fundamental task in clinical research, aiming to estimate the time to an event such as death or relapse. The Cox Proportional Hazards model \cite{cox_og} remains the classical reference due to its interpretability and solid theoretical basis. Regularized extensions, notably LASSO penalization \cite{tibshirani1997}, have enabled its application to high-dimensional settings. More recently, neural-network-based survival models have been proposed to capture non-linear effects and complex feature interactions. DeepSurv \cite{Katzman_2018} adapted the Cox partial likelihood to deep architectures for personalized prediction from structured static data. To model longitudinal dynamics, Dynamic-DeepHit \cite{dynamicdeephit2019} introduced recurrent networks with temporal attention. 
DySurv \cite{dysurv2024} further incorporated conditional variational autoencoders to integrate time-dependent EHR features.


From a mathematical perspective, rough path theory has recently emerged as a principled framework for encoding temporal data. CoxSig \cite{CoxSig} introduced signature transforms into survival modeling, demonstrating strong empirical performance together with theoretical guarantees such as universality and invariance to time reparameterization.

The authors of SigBERT \cite{minchella2025sigbert} introduced a framework combining a domain-specific language model with rough path signature features for temporal survival analysis in oncology. Applied to narrative reports from the Léon Bérard Center, SigBERT achieved a concordance index of 0.75 (sd 0.014) and a time-dependent AUC of 0.80 (sd 0.029), demonstrating that properly encoded clinical text can substantially improve survival risk estimation.

However, SigBERT focuses exclusively on narrative medical reports and does not explicitly address the integration of additional clinical modalities. More broadly, the principled combination of heterogeneous clinical data sources remains a major challenge in survival modeling. A summary of the related methods is provided in the first table of the Supplementary Material.

\subsection{Our Contributions}

Building upon SigBERT \cite{minchella2025sigbert}, which focused on narrative medical reports, we extend this approach to a multimodal setting. Moreover, the original framework was not formulated within a landmark prediction design, which may allow temporal information leakage when constructing temporal features.

To address these limitations, we introduce , a multimodal survival modeling framework designed to integrate heterogeneous clinical data in oncology. The proposed approach jointly leverages unstructured clinical text, structured numerical variables, and their longitudinal evolution within a unified temporal representation.
More precisely, this work makes three main contributions. 

First, we introduce a multimodal temporal encoding framework that combines textual embeddings and structured clinical variables into synchronized trajectories prior to signature transformation. Narrative reports are first encoded into sentence embeddings using OncoBERT \cite{oncobert}. Structured numerical variables are processed separately and aligned with the same temporal axis, thanks to linear interpolation. The two modalities are then concatenated at each time point to form multimodal vectors, which are subsequently encoded using the signature transform, enabling the model to capture higher-order temporal dependencies and cross-modal interactions in a mathematically principled way.

Second, we propose a robust landmark-based survival modeling pipeline that prevents temporal information leakage by restricting feature construction to observations available before the prediction time, thereby enabling well-defined dynamic risk prediction from longitudinal clinical data.

Finally, we provide an efficient and interpretable survival modeling framework combining signature features with sparse Cox regression. This design enables scalable analysis of large clinical cohorts while maintaining interpretability of the predictive factors. In addition, we conduct a systematic empirical comparison of modality-specific contributions by evaluating text-only fully multimodal configurations, thereby quantifying the added value of multimodal fusion for survival prediction.

To our knowledge,  is the first survival framework that jointly combines multimodal temporal encoding of textual and structured numerical data with sparse survival modeling within a coherent and computationally efficient pipeline. 

The complete implementation is available in the associated GitHub repository at 
\texttt{\color{blue}\url{https://github.com/MINCHELLA-Paul/MultiSigBERT}}. 
The main notebook \texttt{multisigbert\_study.ipynb} contains additional experimental results, and all descriptive statistics figures are provided in \linebreak\verb|results/descriptive_statistics|.

\section{Method}

\subsection{Global overview}

The proposed MultiSigBERT framework integrates heterogeneous clinical data into a unified temporal representation for survival modeling. 
The dataset consists of longitudinal patient-level records combining narrative medical reports, structured time-dependent clinical variables, and right-censored survival outcomes.
Formally, for each patient $i$, we observe a sequence of clinical time points 
\begin{equation*}
\left\{ t^{(i)}_{1}, \dots, t^{(i)}_{{N_i}} \right\},
\end{equation*}
at which textual reports and structured covariates are recorded. 
Each report is encoded as a sentence embedding, yielding a time-indexed representation of the patient’s clinical trajectory. 
In addition, we observe structured longitudinal variables measured at irregular time points and aligned with the same temporal axis.

The survival outcome is defined by a pair $(T_i, D_i)$, where 
$T_i \ge 0$ denotes the observed time-to-event or censoring, and 
$D_i \in \{0,1\}$ is the event indicator, with $D_i = 1$ if the event (e.g., death) is observed and $D_i = 0$ if the observation is right-censored. 
The objective is to learn a mapping from the patient-specific temporal representation to a risk function compatible with the Cox proportional hazards framework.

For each patient, clinical notes are observed at successive timestamps, forming an ordered sequence of unstructured text that reflects disease progression, treatment response, and medical decisions over time. In our cohort, these documents consist mainly of consultation reports (68\%) and hospitalization reports (27\%), offering a rich temporal source of prognostic information. In parallel, structured variables such as \textit{Weight}, \textit{Karnofsky Index}, \textit{Blood Pressure}, and \textit{Pulse Rate} are recorded at irregular timepoints. These quantitative measures provide complementary information about the patient’s physiological state. All modalities are preprocessed and temporally aligned to construct coherent multimodal trajectories. Detailed descriptive statistics are reported in the notebook \verb|descriptive_statistics.ipynb| provided in the associated GitHub repository.

To prevent information leakage and define a consistent prediction task, we adopt a landmark design. 
For a fixed landmark time $\boldsymbol{\mathrm{L}}$, only data observed within the backward window $[\boldsymbol{\mathrm{L}}-\boldsymbol{\mathrm{w}}, \boldsymbol{\mathrm{L}}]$ are used to construct the patient representation, and survival is predicted beyond $\boldsymbol{\mathrm{L}}$. 
This framework preserves temporal causality and ensures that signature coefficients are computed on finite, well-ordered trajectories without incorporating future information.

Each modality is first embedded in a dedicated vector space, optionally compressed, and concatenated across time. The resulting multimodal sequences are encoded using the signature transform, producing fixed-dimensional representations that capture higher-order temporal interactions.
Moreover, the signature transform is particularly well suited to handle irregularly sampled time points and heterogeneous data modalities, both of which constitute major challenges in real-world machine learning settings.

Finally, these signature-derived features are used to fit a LASSO-regularized Cox model, producing individualized risk scores while ensuring sparsity, strong predictive performance, and computational efficiency.

\begin{figure}[H]
\centering
\includegraphics[width=\textwidth]{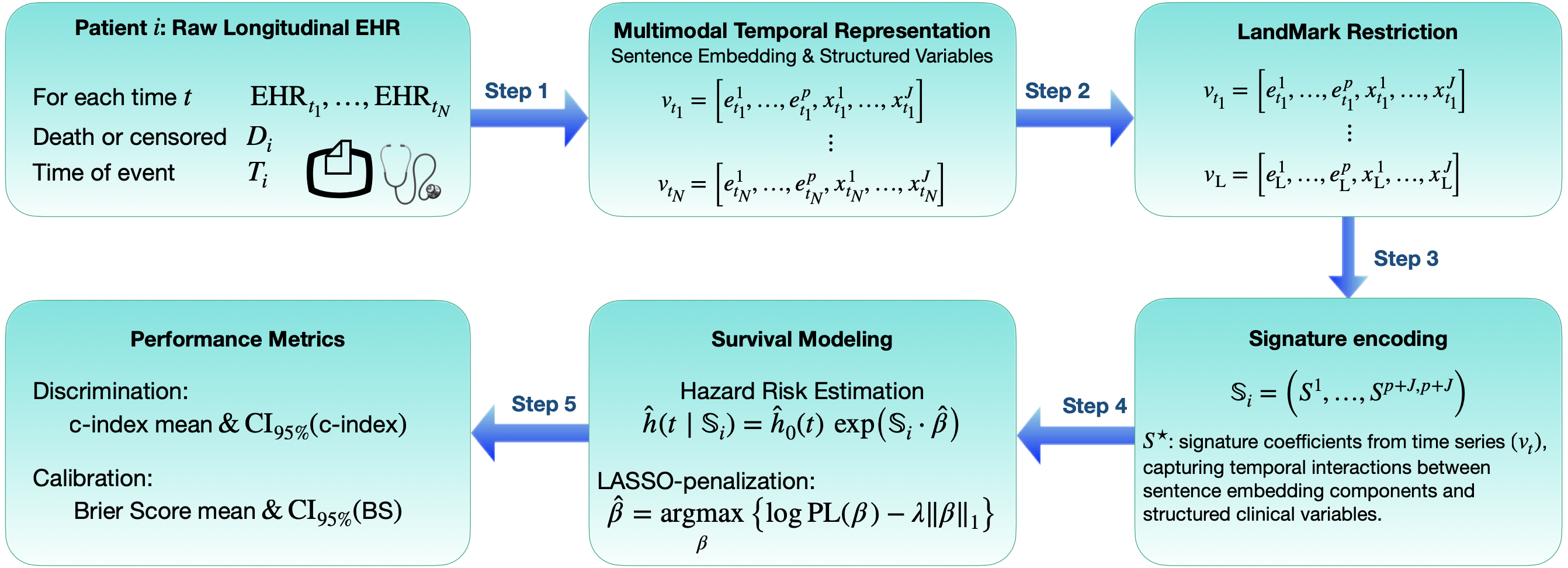}
\caption{
\textbf{Overview of the MultiSigBERT pipeline.}
\textbf{\textit{Step 1: Multimodal representation.}}
Narrative clinical reports are converted into sentence embeddings and combined with structured variables observed at times $t_1,\dots,t_N$ to form multimodal vectors $v_t$.
\textbf{\textit{Step 2: Landmark restriction.}}
Only observations within the window $[\boldsymbol{\mathrm{L}}-\boldsymbol{\mathrm{w}}, \boldsymbol{\mathrm{L}}]$ are retained to avoid temporal information leakage.
\textbf{\textit{Step 3: Signature encoding.}}
The truncated sequence $(v_t)_t$ is encoded via the path signature transform, producing a fixed-dimensional representation $\mathbb{S}_i$ capturing temporal interactions across modalities.
\textbf{\textit{Step 4: Survival modeling.}}
A LASSO-regularized Cox model is fitted on the signature features to estimate individualized risk scores.
\textbf{\textit{Step 5: Evaluation.}}
Performance is assessed using the C-index and the Brier score.
}
\label{fig:pipeline_multimodal}
\end{figure}

An overview of the complete pipeline -- from embedding extraction and temporal encoding to geometric clustering and survival estimation -- is illustrated in Figure~\ref{fig:pipeline_multimodal}.

\subsection{Multimodal Embedding Extraction}

Clinical datasets combine structured numerical variables with unstructured data such as free-text reports. To integrate these heterogeneous modalities into a unified survival framework, each input must first be transformed into a numerical vector representation.

Let's first consider narrative medical reports. These longitudinal documents capture the patient’s clinical follow-up, including consultation reports, hospitalization summaries, and medical observations recorded over time. We encode them using OncoBERT \cite{oncobert}, a CamemBERT-based \cite{martin2020camembert} language model fine-tuned on oncology notes from the Léon Bérard Center. This domain-specific fine-tuning improves the semantic representation of oncology terminology and clinical context.
Let $\mathscr{R}_i(t)$ denote the raw clinical report associated with patient $i$ at time $t$. The OncoBERT encoder
$
\Phi_{\mathrm{OncoBERT}}
$
maps this report into a sequence of contextualized token embeddings:
\[
\big(e_w\big)_{w\in\mathscr{R}_i(t)} = \Phi_{\mathrm{OncoBERT}}(\mathscr{R}_i(t)),
\]
where $e_w\in\mathbb{R}^{p}$ denotes the contextualized embedding associated with token $w$, with $p=768$.
A fixed-dimensional sentence representation
$
v_i(t)\in\mathbb{R}^{p}
$
is then obtained from these contextualized token embeddings using the Smooth Inverse Frequency (SIF) pooling strategy \cite{arora2017asimple}. Specifically,
\begin{equation*}
v_i(t)
=
\frac{1}{|\mathscr{R}_i(t)|}
\sum_{w\in\mathscr{R}_i(t)}
\frac{a}{\mathrm{f}(w)+a}
\,e_w,
\end{equation*}
where $\mathrm{f}(w)$ denotes the frequency of word $w$, $|\mathscr{R}_i(t)|$ denotes the number of tokens in the report, and $a$ is a smoothing parameter, typically set to $a=10^{-3}$.

The resulting vector $v_i(t)$ constitutes the sentence embedding associated with patient $i$ at time $t$ and serves as the elementary observation in the subsequent longitudinal trajectory. 
We adopt Smooth Inverse Frequency (SIF) embeddings rather than the standard CLS token representation. 
Empirically, SIF consistently outperformed CLS in our retrospective experiments, yielding higher C-index values (0.74 vs.\ 0.70) and lower Brier scores. 
This gain can be attributed to the re-weighting and averaging mechanism of SIF, which mitigates the influence of frequent or uninformative tokens and produces more stable sentence representations. 
In addition, SIF naturally accommodates long clinical documents by aggregating token embeddings beyond the 512-token limitation of BERT models. 
For a detailed methodological and empirical discussion, we refer the reader to~\cite{minchella2025sigbert}.
A patient with $N_i$ reports is thus represented by the time-indexed sequence of sentence embeddings
\[
\big(v_i(t_1), \dots, v_i(t_{N_i})\big),
\qquad
v_i(t_j)\in\mathbb{R}^{p},
\]
which can equivalently be viewed as an element of $\mathbb{R}^{N_i\times p}$.

Since the signature transform scales exponentially with the input dimension, dimensionality reduction is required. We apply Principal Component Analysis (PCA) on the training embeddings and retain the first $r$ principal components, yielding a projection matrix
$
R_{\mathrm{comp}}\in\mathbb{R}^{r\times p}.
$
Each sentence embedding is then compressed according to
\[
\widetilde{v}_i(t)
=
R_{\mathrm{comp}}\cdot v_i(t)
\in\mathbb{R}^{r}.
\]
The resulting trajectory
$
\big(\widetilde{v}_i(t_1),\dots,\widetilde{v}_i(t_{N_i})\big)
$
is subsequently used for temporal encoding through the signature transform.
The dimension $r$ is selected empirically to balance predictive performance and computational cost. In our experiments, performance stabilized from $r=25$, which we then adopt in the sequel.

\subsection{Integration of Structured Clinical Variables}

Beyond text, patients are characterized by longitudinal structured variables being \textit{Weight}, \textit{Karnofsky Index}, \textit{Blood Pressure}, and \textit{Pulse} (meaning that the number of structured clinical covariates is given by $K=4$). These measurements form irregular time series that reflect physiological evolution.
To integrate them with textual embeddings, we construct synchronized multimodal trajectories. For each structured variable $x^{k}$, $k=1,\dots,K$, observations are first sorted chronologically; missing internal values are then linearly interpolated, boundary values are propagated forward or backward when necessary, and all variables are finally standardized to zero mean and unit variance.
At each timestamp $t$, the compressed textual representation of patient $i$ is given by
$\widetilde{v}_i(t)\in\mathbb{R}^{r}$, 
while the structured measurements are represented by
\[
x_i(t)
=
\Big(
x_i^{1}(t),
\dots,
x_i^{K}(t)
\Big)
\in\mathbb{R}^{K}.
\]
In our setting, a patient $i$ will be represented by the synchronized multimodal trajectory
\[
X_i : [0,T_i] \longrightarrow \mathbb{R}^{q},
\]
whose coordinates define a collection of $q$ one-dimensional paths evolving over time. The ambient dimension $q$ is obtained by concatenating the compressed embedding representation of size $r$, the $K$ structured clinical variables, and the temporal coordinate required to preserve the chronological ordering of observations, yielding
$
q=r+K+1.
$ The multimodal observation at time $t$ is defined by concatenation:
\begin{equation*}
\label{eq:time-series-patient}
X_i(t)
=
\Big(
t,\widetilde{v}_i(t),
x_i^{1}(t),
\dots,
x_i^{K}(t)
\Big)
\in\mathbb{R}^{q},
\end{equation*}
where the first coordinate corresponds to the temporal component, $\widetilde{v}_i(t)\in\mathbb{R}^r$ denotes the compressed textual representation extracted from the clinical report, and $x_i^k(t)$ denotes the value of the $k$-th structured clinical variable at time $t$.
Given the observation times
$
0 \le t_1 < \cdots < t_{N_i} \le T_i,
$
the trajectory of patient $i$ can be represented equivalently as the matrix
\begin{equation}
\label{eq:temp-covar-patient}
X_i = \big( X_i(t_1), \dots, X_i(t_{N_i}) \big)^\top \in \mathbb{R}^{N_i\times q},
\end{equation}
whose rows correspond to successive multimodal observations and whose columns correspond to the different coordinates of the trajectory.
This representation jointly encodes the semantic evolution extracted from clinical narratives and the temporal evolution of structured physiological measurements over the $N_i$ available observation times.

In the next section, the signature transform is applied to the trajectory $X_i$ in order to extract higher-order temporal features describing the patient's longitudinal evolution.

\subsection{Signature Feature Extraction}

The signature of a path \cite{chen1954}, later adapted to rough path theory by Lyons \cite{lyons1998differential}, provides a systematic way of encoding sequential data through collections of iterated integrals.
To ensure uniqueness of the signature representation, a monotone coordinate is typically appended to the path, most commonly the time component. 
Writing
$
X_i(t)
=
\bigl(
X_i^{1}(t),
\dots,
X_i^{q}(t)
\bigr),
$
the iterated-integral signature coordinate associated with the multi-index
$
(j_1,\dots,j_\ell)\in\{1,\dots,q\}^{\ell}
$
is defined by
\begin{equation*}
\label{sign_def}
S(X_i)_{0,t}^{(j_1,\dots,j_\ell)} = \int_{0<t_1<\cdots<t_\ell<t} \mathrm{d}X_i^{j_1}(t_1) \cdots \mathrm{d}X_i^{j_\ell}(t_\ell).
\end{equation*}

The collection of these features is organized in tensor form which uniquely encodes the path and is defined as: 
\begin{equation*}
    S^{\ell}(v) = \Big( S(v)_{0, T}^{(j_{1}, \ldots, j_{\ell})} \Big)_{(j_{1}, \ldots, j_{\ell}) \in \{ 1, \dots, q\}^\ell} \in (\mathbb{R}^q)^{\otimes \ell}.
\end{equation*}
Thus, the truncated signature up to order $L$ naturally belongs to the truncated tensor algebra $\mathcal{T}^{\leq L}(\mathbb{R}^q) = \bigoplus\limits_{\ell=0}^{L} \, (\mathbb{R}^q)^{\otimes \ell}$ of order $L$ over $\mathbb{R}^q$ :
\begin{equation*}
    S^{\leq L}(v) := \big( S^{\ell}(v) \big)_{\ell=0}^{L} \in \mathcal{T}^{\leq L}(\mathbb{R}^p).
\end{equation*}

In the remainder of this work, we explicitly denote the truncated-signature operator of order $L$ by
\[
\operatorname{Sig}^{(L)}(X_i)
:=
S^{\leq L}(X_i),
\]
denoting the collection of all iterated-integral coordinates of the multimodal trajectory $X_i$ up to level $L$.

In addition to encoding temporal dynamics, this approach handles sequences of varying lengths and is invariant to translation and temporal reparameterization (see \cite{Chevyrev2016}), making it well-suited for patients with different study entry points and durations. 
Finally, for a given patient $i$, the collection of signature covariates extracted from the multimodal trajectory $X_i$, truncated at order $m=2$, is denoted by
$\mathbb{S}_i=\operatorname{Sig}^{(2)}(X_i)$.
More explicitly, the truncated signature can be written as
\begin{equation*}
\label{sig_covariates}
\mathbb{S}_i
=
\Big(
1,\,
S^{(1)},\ldots,S^{(q)},\,
S^{(1,1)},\ldots,S^{(q,q)}
\Big)_i.
\end{equation*}
Each patient trajectory, initially represented as a time series, is therefore summarized by a fixed-dimensional vector of signature coefficients $\mathbb{S}_i$ encoding its temporal dynamics. These signature covariates provide a structured representation of longitudinal evolution in a Euclidean space. This representation avoids the direct manipulation of irregular time series and enables the use of standard machine learning models for downstream tasks while preserving temporal consistency.

\subsection{Survival Analysis Modeling}

The Cox proportional hazards model \cite{cox_og} remains the gold standard in survival analysis, notably because it naturally accounts for right-censored observations, i.e., patients for whom the event of interest $T$ (such as death or relapse) has not yet occurred during the observation period. In this work, we therefore adopt the Cox model to illustrate the impact of combining textual data with clinical variables.
All censored patients still contribute to the likelihood estimation, helping to reduce bias and improve the robustness of predictions. The goal is to estimate the probability of a patient surviving beyond time $t$, noted as $\mathcal{S}(t\mid \mathbb{S}) := \mathbb{P} \big( T \geq t \mid \mathbb{S} \big)$ when knowing their covariates $\mathbb{S}$. This estimation relies on the key concept of instantaneous hazard rate $h$, which quantifies the infinitesimal probability of the event occurring at $t$ and is related to survival through the following equation: \begin{equation*}
    \label{ODE_risk_surv}
    \mathcal{S}(t\mid \mathbb{S}) = \exp\left( - \int^{t} h(s\mid \mathbb{S}) \, \mathrm{d}s\right).
\end{equation*} David R. Cox proposed the generalized linear model: \begin{equation*}
    \label{cox_hazard_model}
    h(t\mid \mathbb{S}) \, = \, h_0(t) \cdot \exp\big( \mathbb{S} \cdot \boldsymbol{\beta} \big)
, \end{equation*} where $\boldsymbol{\beta} \in \mathbb{R}^{Q}$ is the vector of parameters to be estimated, and $h_0$ is the baseline hazard, common to all patients, as estimated by \cite{breslow1972}. We define $\eta := \mathbb{S} \cdot \boldsymbol{\beta}$, referred to as the risk score. Estimating $\boldsymbol{\beta}$ involves managing a substantial number of covariates. As mentioned earlier, this is due to the signature transform, which generates a high-dimensional feature space: for $q$ input channels and a truncation level $L$, the number of resulting signature coefficients scales as $O(q^L)$. Even after dimensionality reduction, the resulting covariate space remains large. To reduce the risk of overfitting and improve model stability, we apply the LASSO (\textit{Least Absolute Shrinkage and Selection Operator}) regularization to the Cox model, as originally introduced by Robert Tibshirani in \cite{tibshirani1997}: \begin{equation}
    \label{lasso_pen}
    \widehat{\boldsymbol{\beta}} \in \underset{\boldsymbol{\beta}}{\operatorname{argmax}} \ \log \text{PL}(\boldsymbol{\beta}) - \lambda \| \boldsymbol{\beta} \|_1,
\end{equation} where $\text{PL}(\boldsymbol{\beta})$ is the partial likelihood defined and fully detailed in \cite{partial_likelihood}, and $\lambda > 0$ denotes the regularization parameter. $ \|\boldsymbol{\beta}\|_1 $ is the $\ell_1$-norm of the parameters $\boldsymbol{\beta}$. The impact of LASSO regularization is twofold: it shrinks some coefficients towards zero, effectively removing less relevant covariates, and it selects only the most important predictors for survival, enhancing model stability. 


Beyond its statistical advantages, this sparsity pattern provides structural insight: only a limited subset of signature coordinates appears to carry prognostic information. In other words, although the signature transform embeds patient trajectories into a high-dimensional feature space, the effective predictive signal lies along a small number of directions. This suggests that progression patterns may have a low intrinsic dimensionality within the signature representation, highlighting the importance of both regularization and structured feature extraction in our framework. 
Moreover, by enforcing sparsity, the LASSO-regularized Cox model significantly reduces the number of active covariates, leading to faster computational performance. This suggests that the model achieves a favorable balance between overfitting and underfitting, leveraging a compact and efficient representation of the risk factors while maintaining strong predictive power.

To put this idea into practice, model parameters are obtained by maximizing the penalized objective function in \eqref{lasso_pen}, whose explicit, form when $\log$ applied, is given by:
\begin{equation*}
    \label{objective_fct_lasso}
    \mathcal{L}(\boldsymbol{\beta}) = \sum_{i: \delta_i = 1} \left[ \mathbb{S}_i \boldsymbol{\beta} - \log  \sum_{j \in \mathcal{R}_i} \exp\Big(\mathbb{S}_j \boldsymbol{\beta}\Big)  \right] - \lambda \sum_{k=1}^{Q} |\beta_k | .
\end{equation*}
Here, $\delta_i \equiv \mathbbm{1}_{\{D_i=1\}} \in \{0,1\}$ indicates whether the event (e.g., death) has been observed for patient $i$, with associated study duration $T_i$. The set $\mathcal{R}_i$ denotes the risk set, i.e., the individuals still at risk at time $T_i$, formally defined as $\mathcal{R}_i = \{ j : T_j \geq T_i \}$.

Finally, the estimated risk score under LASSO regularization is obtained simply as the dot product $$    \widehat{\eta} = \mathbb{S} \cdot \widehat{\boldsymbol{\beta}}.$$ Each patient characterized by a sequence of clinical reports and associated sequential covariates is assigned an individualized risk score $\widehat{\eta}$ that summarizes the temporal evolution of their clinical trajectory. Overall, this pipeline provides a straightforward framework to integrate heterogeneous EHR modalities into a unified representation suitable for survival analysis.

To ensure a well-defined prediction task and avoid temporal information leakage, it is however necessary to construct these representations using only information available up to a given prediction time. In the next section, we therefore introduce a landmark design that restricts the computation of signature coefficients to observations occurring before a fixed time point, ensuring that the model is trained exclusively on past information.

\subsection{Landmark Modeling for Dynamic Survival Prediction}

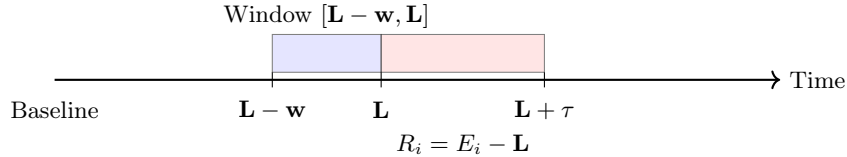
\begin{figure}
\centering
\begin{tikzpicture}[xscale=0.12,yscale=1]

\draw[->, thick] (0,0) -- (80,0) node[right]{\footnotesize Time};
\node at (0,-0.4) {\footnotesize Baseline};

\draw[fill=blue!20,opacity=0.5] (24,0.1) rectangle (36,0.6);
\node at (30,0.85) {\footnotesize Window $[\boldsymbol{\mathrm{L}}-\boldsymbol{\mathrm{w}},\boldsymbol{\mathrm{L}}]$};

\draw[fill=red!20,opacity=0.5] (36,0.1) rectangle (54,0.6);
\node at (45,-0.85) {\footnotesize $R_i=E_i-\boldsymbol{\mathrm{L}}$};

\foreach \x/\lab in {24/{$\boldsymbol{\mathrm{L}}-\boldsymbol{\mathrm{w}}$},36/{$\boldsymbol{\mathrm{L}}$},54/{$\boldsymbol{\mathrm{L}}+\tau$}} {
  \draw (\x,0.1)--(\x,-0.1);
  \node at (\x,-0.4) {\footnotesize \lab};
}

\end{tikzpicture}
\caption{Landmark design: features are constructed from the backward window $[\boldsymbol{\mathrm{L}}-\boldsymbol{\mathrm{w}},\boldsymbol{\mathrm{L}}]$ and survival is re-indexed by $R_i=E_i-\boldsymbol{\mathrm{L}}$. The prediction task is to estimate the conditional survival probability $\mathbb{P}(T_i>\boldsymbol{\mathrm{L}}+\tau \mid \mathbb{X}_i(\boldsymbol{\mathrm{L}}))$, using only information observed before the landmark time.}
\label{fig:landmark_tikz}
\end{figure}

Landmarking provides a principled framework for dynamic survival prediction by conditioning on a fixed prediction time $\boldsymbol{\mathrm{L}}$ \cite{anderson1983landmark,vanHouwelingen2008}. Rather than using the entire longitudinal process, including information recorded after the prediction time, the analysis is restricted to the landmark risk set
\begin{equation}
\label{eq:cohort-landmark}
\mathcal{R}_{\boldsymbol{\mathrm{L}}}
=
\bigl\{
i : E_i \ge \boldsymbol{\mathrm{L}}
\bigr\},
\end{equation}
where
$
E_i=\min(T_i,C_i)
$
denotes the observed survival time, obtained from the true event time $T_i$ and censoring time $C_i$. Consequently, only patients who remain under observation at time $\boldsymbol{\mathrm{L}}$ contribute to the landmark analysis.
For each patient $i\in\mathcal{R}_{\boldsymbol{\mathrm{L}}}$, survival is re-indexed relative to the landmark:
\[
R_i
=
E_i-\boldsymbol{\mathrm{L}}
\ge 0.
\]
The corresponding landmark event indicator is defined by
$
\delta_i(\boldsymbol{\mathrm{L}})
=
\delta_i
\,
\mathbbm{1}_{\{E_i>\boldsymbol{\mathrm{L}}\}}
$
where
$
\delta_i
=
\mathbbm{1}_{\{T_i\le C_i\}}
$
still denotes the usual event indicator. This formulation ensures that both the outcome and the covariates are defined relative to the same prediction time $\boldsymbol{\mathrm{L}}$, thereby preventing temporal leakage and guaranteeing that predictions are conditional on survival up to the landmark.
An illustration of the landmark principle is provided in Figure~\ref{fig:landmark_tikz}.

Let's apply this setup to our context. For a fixed landmark time $\boldsymbol{\mathrm{L}}$, feature extraction is restricted to the backward window
$
[\boldsymbol{\mathrm{L}}-\boldsymbol{\mathrm{w}},
\boldsymbol{\mathrm{L}}],
$
thereby ensuring that only information available prior to the prediction time is used. Denoting by
$
X_i {(\boldsymbol{\mathrm{L}})}
$
the restriction of the trajectory $X_i$ (defined in~\ref{eq:temp-covar-patient}) to this interval, the landmark-specific signature representation is defined as
\[
\mathbb{S}_i(\boldsymbol{\mathrm{L}})
=
\operatorname{Sig}^{(2)}
\!\left(
X_i {(\boldsymbol{\mathrm{L}})}
\right).
\]

To account for patients with limited clinical history prior to the landmark, we introduce the binary indicator
\[
\zeta_i(\boldsymbol{\mathrm{L}})
=
\mathbbm{1}_{\left\{
\boldsymbol{\mathrm{L}}
-
t_i^{\mathrm{diag}}
<
\boldsymbol{\mathrm{w}}
\right\}},
\]
where $t_i^{\mathrm{diag}}$ denotes the diagnosis time (or, equivalently, the first available clinical observation). This indicator identifies patients whose available history before $\boldsymbol{\mathrm{L}}$ is shorter than the landmark window. This indicator is particularly relevant in landmark modeling, as some patients may have entered the study less than $\boldsymbol{\mathrm{w}}$ time units before the landmark and therefore lack a complete history over the window $[\boldsymbol{\mathrm{L}}-\boldsymbol{\mathrm{w}},\boldsymbol{\mathrm{L}}]$.
The final landmark covariate vector is obtained by concatenation:
\[
\mathbb{X}_i(\boldsymbol{\mathrm{L}})
=
[\mathbb{S}_i(\boldsymbol{\mathrm{L}}), \ 
\zeta_i(\boldsymbol{\mathrm{L}})]^\top \in \mathbb R ^{Q+1}.
\]
The resulting landmark dataset consists of triplets
$
\Bigl(
\mathbb{X}_i(\boldsymbol{\mathrm{L}}),
\,
\delta_i(\boldsymbol{\mathrm{L}}),
\,
R_i
\Bigr),
$ for each $i\in\mathcal{R}_{\boldsymbol{\mathrm{L}}}$,
which are used to estimate conditional survival beyond the landmark time.
More precisely, the prediction task consists of estimating
\[
\mathbb{P}
\Bigl(
T_i
>
\boldsymbol{\mathrm{L}}
+
\tau
\;\Big|\;
\mathbb{X}_i(\boldsymbol{\mathrm{L}})
\Bigr),
\]
that is, the probability that patient $i$ survives an additional horizon $\tau$ beyond the landmark time, given all information available up to $\boldsymbol{\mathrm{L}}$.

This framework is particularly well suited to longitudinal clinical data and signature-based representations. By construction, it ensures temporal coherence, prevents information leakage, accommodates irregular follow-up patterns, and allows patients with heterogeneous observation histories to be compared at a common prediction time.

\section{Experiments}

\subsection{Cohort}

This study complies with the General Data Protection Regulation (GDPR) and falls within the scope of scientific research conducted in the legitimate interest of cancer research, in accordance with Articles 6.1.f and 9.2.j of Regulation (EU) No. 2016/679. This project has been officially registered under the MR004 declaration (V3.2, 23/08/2021) at the Léon Bérard Center, ensuring compliance with legal and ethical standards for processing health data. The data have been carefully anonymized and {can only be used within the framework of this study}. No patient were opposed to this study. To ensure the reliability of our data, we selected a study cohort consisting of patients hospitalized -- at least once -- at the Léon Bérard Center from 2000 to 2024, with comprehensive follow-up throughout their medical care to ensure data completeness and accuracy. 

All results reported in this section are obtained under a landmark design with landmark time $\boldsymbol{\mathrm{L}}=36$ months and backward window size $\boldsymbol{\mathrm{w}}=12$ months. In other words, we consider patients who remain at risk three years after their initial diagnosis or first hospitalization, and construct covariates using information collected during the one-year period preceding the landmark.

The dataset consists of longitudinal medical reports collected from a comprehensive oncology cohort. It includes {2,527} patients and a total of {124,049} clinical reports, with an average of {49} reports per patient, reflecting dense longitudinal follow-up. Among these patients, {1,629} experienced the event of interest and {898} were right-censored.

The cohort covers a wide spectrum of cancer types. The most represented diagnoses are breast cancer ({782 patients}, 31.0\%), and to name a few more, gynecological cancers ({232}, 9.2\%), intestinal cancers ({227}, 9.0\%), prostate cancer ({201}, 8.0\%), and lung cancer ({96}, 3.8\%). Additional cases include other oncological subtypes, ensuring heterogeneity in disease trajectories and survival patterns.
The longitudinal structure of the data makes it particularly suitable for landmark modeling. At each landmark time $\boldsymbol{\mathrm{L}}$, only patients still at risk are considered, and covariates are constructed from the backward window $[\boldsymbol{\mathrm{L}}-\boldsymbol{\mathrm{w}},\boldsymbol{\mathrm{L}}]$, ensuring temporal coherence and preventing information leakage.
Descriptive statistics and additional cohort characteristics (cancer types, number of reports per patient, study duration, diagnosis dates, and Kaplan--Meier curves) are available in the associated GitHub repository (\texttt{\color{blue}\url{https://github.com/MINCHELLA-Paul/MultiSigBERT/tree/main/results/descriptive\_statistics}}).

\subsection{Hyperparameter Search}

Our experimental setup is designed to ensure reproducibility, robustness, and realistic evaluation in a complex real-world clinical context. The NLP model used in this study OncoBERT is previously fine-tuned on oncology-specific clinical notes. It is used as is throughout the experiments, without further task-specific adaptation, in order to reflect realistic deployment scenarios. The survival model was trained on a cohort of $1,264$ patients ($9,696$ reports) and evaluated on a separate test set of $1,263$ patients ($9,752$ reports), using a structured and stratified train-test split to preserve temporal and distributional consistency. To calibrate the model, we conducted a grid search for the LASSO regularization parameter $\lambda$ within the range $[0.01, 16]$, using a fixed step size of $0.01$. We selected the value that maximized the cross-validation concordance index (C-index) averaged over five independent validation folds, within the training dataset. 

At this step, the longitudinal data were transformed using the signature method, effectively eliminating any temporal constraints that could arise when subdividing the dataset for cross-validation. The selection criterion aimed to maximize the C-index through five-fold cross-validation. Specifically, for each candidate value of $\lambda$, the model was trained on a partition of the training cohort and evaluated on held-out subsets -- within the training set --, maintaining equal proportions across folds.

The optimal value was then chosen as the one yielding the highest mean C-index, promoting robust generalization and avoiding overfitting. This tuning process was crucial for balancing model sparsity and predictive accuracy.

\begin{table}[H]
\centering
\renewcommand{\arraystretch}{1.1} 
\setlength{\tabcolsep}{5pt} 

\begin{tabular}{|l||c|c|c|}
\toprule
\textbf{Scenario} & $\boldsymbol{\lambda}$ & \textbf{C-index Val.} $\uparrow$ & \textbf{IBS@3y} $\downarrow$ \\
\midrule
Reports Only 
& $5.15$ 
& $0.747\;(\mathrm{sd}\;0.027)$ 
& $0.1280\;(\mathrm{sd}\;0.004)$ \\

Reports \& Structured Variables 
& $4.60$ 
& $0.743\;(\mathrm{sd}\;0.025)$ 
& $0.1280\;(\mathrm{sd}\;0.004)$ \\


\bottomrule
\end{tabular}
\vspace{5pt}
\caption{Optimal regularization LASSO-parameter $\lambda$ selected by validation for each modeling scenario. 
Performance is reported as the mean C-index on the validation sets with its standard deviation 
(higher is better) and the Integrated Brier Score at 3 years (IBS@3y, lower is better).}
\label{tab:lambda_selection_results}
\end{table}

\subsection{Performance Metrics}

Model validation is performed using repeated random splitting: the test set is divided into ten disjoint subsets, and evaluation is repeated independently on each split. Reported values correspond to the mean and standard deviation across splits.

We first assess discrimination using the concordance index (C-index) \cite{harrell1982evaluating}, computed on the landmark cohort
$
\mathcal{R}_{\boldsymbol{\mathrm{L}}}
$ from~\eqref{eq:cohort-landmark}.
The C-index measures the proportion of correctly ordered comparable pairs. If patient $j$ experiences the event before patient $i$, a well-specified model should assign a higher risk score $\widehat{\eta}_j > \widehat{\eta}_i$. Using the landmark outcomes $(R_i,\delta_i(\boldsymbol{\mathrm{L}}))$, it can be written as
\[
\mathrm{C}\text{-}\mathrm{index} = \frac{ \sum\limits_{i,j\in\mathcal{R}_{\boldsymbol{\mathrm{L}}}} \mathbbm{1}_{\{R_j<R_i\}} \mathbbm{1}_{\{\widehat{\eta}_j>\widehat{\eta}_i\}} \delta_j(\boldsymbol{\mathrm{L}}) }{ \sum\limits_{i,j\in\mathcal{R}_{\boldsymbol{\mathrm{L}}}} \mathbbm{1}_{\{R_j<R_i\}} \delta_j(\boldsymbol{\mathrm{L}}) }.
\]
A value of $1$ indicates perfect ranking, whereas $0.5$ corresponds to random performance.

While discrimination remains important, calibration has become equally critical in modern survival modeling. A model may rank patients correctly while providing inaccurate survival probabilities. To assess calibration, we use the Brier Score (BS), which measures the squared difference between the predicted landmark survival probability
$
\widehat{\mathcal S}(t\mid \mathbb X_i(\boldsymbol{\mathrm L}))
$
and the observed landmark outcome. To account for right censoring, we employ inverse probability of censoring weighting (IPCW) \cite{graf1999brier}:
\begin{equation}
\label{eq:BS}
\mathrm{BS}(t)
=
\frac{1}{|\mathcal R_{\boldsymbol{\mathrm L}}|}
\sum_{i\in\mathcal R_{\boldsymbol{\mathrm L}}}
\Bigg[
\mathbbm{1}_{\{R_i\le t,\ \delta_i(\boldsymbol{\mathrm L})=1\}}
\frac{
\widehat{\mathcal S}(t\mid\mathbb X_i(\boldsymbol{\mathrm L}))^2
}
{\widehat G(R_i)}
+
\mathbbm{1}_{\{R_i>t\}}
\frac{
\Big(
1-\widehat{\mathcal S}(t\mid\mathbb X_i(\boldsymbol{\mathrm L}))
\Big)^2
}
{\widehat G(t)}
\Bigg].
\end{equation}
where $\widehat G$ denotes the Kaplan--Meier estimator \cite{KaplanMeier1958} of the censoring distribution. In the landmark setting, the Brier Score evaluates the accuracy of predicted residual survival probabilities beyond the landmark time $\boldsymbol{\mathrm L}$. 

The Brier Score directly quantifies the accuracy of predicted survival probabilities. This property is crucial in clinical applications, where predicted survival probabilities may influence treatment decisions, patient stratification, and follow-up strategies. A value substantially below the naive reference value of $0.25$ -- corresponding to a constant prediction of $0.5$ for all individuals -- indicates good calibration. 
Finally, to summarize predictive accuracy over a given time interval rather than at a single time point, we use the Integrated Brier Score (IBS) that provides a global calibration measure by integrating~\eqref{eq:BS} over a time interval $[\tau_1,\tau_2]$. In practice, it can be evaluated up to a fixed horizon, for example 3 years, by setting $\tau_1 = 0$ and $\tau_2 = 3\text{y}$.

\subsection{Experimental Results}

All results obtained under the landmark design ($\boldsymbol{\mathrm{L}}=36$, $\boldsymbol{\mathrm{w}}=12$) are summarized in Table~\ref{tab:results_landmark_main}.
The models adapted to this architecture include DeepSurv~\cite{Katzman_2018}, CoxTime~\cite{kvamme2019coxtime}, and the time-varying Cox model implemented via CoxTimeVaryingFitter~\cite{lifelines}, denoted CTVF. CoxTime and CTVF are trained and applied directly on the longitudinal time-series representation via Equation~\eqref{eq:time-series-patient}, whereas DeepSurv operates on the fixed-length feature representation obtained after computing the signature coefficients.
We consider two experimental configurations: 
(i) \emph{Reports Only}, relying exclusively on signature features extracted from textual embeddings (therefore, corresponding to the initial model SigBERT \cite{minchella2025sigbert}); and 
(ii) \emph{Reports + Sequential Structured Data}, where interpolated longitudinal clinical variables are concatenated with textual embeddings prior to the signature transformation.

Overall, as shown in Table~\ref{tab:results_landmark_main}, MultiSigBERT achieves strong predictive performance under the landmark design. 
In the \emph{Reports Only} configuration, it reaches a mean test C-index of \textbf{0.741} (sd 0.027), substantially outperforming several classical survival baselines such as CoxTime (0.646) and CTVF (0.619). 
It also clearly improves upon neural survival models based on aggregated representations, such as DeepSurv, which attains a mean C-index of 0.600. 
In terms of calibration, MultiSigBERT yields a relatively low prediction error with an $\mathrm{IBS}_{@3y}$ of \textbf{0.128} (sd 0.006), indicating accurate survival probability estimates across the evaluation horizon.

\begin{table}
\centering
\renewcommand{\arraystretch}{1.1}
\setlength{\tabcolsep}{5pt}
\small
\begin{tabular}{|l|c|c|}
\hline
\textbf{Model} & \textbf{C-index $\uparrow$ (Test)} & \textbf{IBS $\downarrow$ (Test)} \\
\hline

\textbf{MultiSigBERT (RO)}
& \textbf{0.741 \;\; [0.724, 0.758]}
& {0.1282 \;\; [0.1243, 0.1322]} \\

DeepSurv \hspace{1cm} (RO)
& {0.600 \;\; [0.583, 0.614]}
& {0.1512 \;\; [0.1406, 0.1619]} \\

CoxTime \hspace{1.07cm} (RO)
& 0.646 \;\; [0.638, 0.654]
& \textbf{0.091 \;\; [0.081, 0.101]} \\

CTVF \hspace{1.55cm}(RO)
& 0.619 \;\; [0.594, 0.645]
& 0.207 \;\; [0.185, 0.229] \\

\hline

\textbf{MultiSigBERT (RS)}
& \textbf{0.743 \;\; [0.725, 0.761]}
& {0.1339 \;\; [0.1293, 0.1385]} \\

DeepSurv \hspace{1.11cm}(RS)
& {0.650 \;\; [0.620, 0.676]}
& {0.1490 \;\; [0.1386, 0.1593]} \\

CoxTime \hspace{1.19cm}(RS)
& 0.651 \;\; [0.640, 0.661]
& \textbf{0.091 \;\; [0.081, 0.101]} \\

CTVF \hspace{1.56cm}(RS)
& 0.618 \;\; [0.594, 0.645]
& 0.209 \;\; [0.185, 0.229] \\

\hline
\end{tabular}

\caption{Test-set performance under the landmark design (Train = 1,263 patients; Test = 1,264 patients). 
Values are reported as mean with 95\% Jackknife confidence intervals in brackets. 
RO: Reports Only; RS: Reports + Sequential Structured Data. 
$\uparrow$ indicates that higher values are better; $\downarrow$ indicates that lower values are better.}

\label{tab:results_landmark_main}
\end{table}

When sequential structured covariates are incorporated, MultiSigBERT maintains essentially the same level of predictive performance. 
The mean test C-index slightly increases to \textbf{0.743} (sd 0.029), while the calibration error becomes $\mathrm{IBS}_{@3y} = 0.134$ (sd 0.007). 
These results suggest that adding structured variables provides only marginal gains in discrimination while slightly degrading calibration.

Compared with neural survival baselines such as DeepSurv, MultiSigBERT offers a substantially better overall trade-off between discrimination and calibration. 
Even when structured variables are included, DeepSurv reaches a mean C-index of only 0.650, remaining clearly below the performance achieved by MultiSigBERT. 
Although CoxTime achieves the lowest $\mathrm{IBS}_{@3y}$ (around 0.091), its discrimination remains substantially lower (C-index $\approx 0.65$), which limits its usefulness for clinical risk stratification where reliable patient ranking is essential.

From a clinical perspective, this trade-off is particularly important. 
Effective decision support requires models that can both correctly rank patients by risk and provide reasonably calibrated survival probabilities. 
In this respect, MultiSigBERT provides a robust compromise: it achieves substantially higher discrimination than traditional Cox-based approaches while maintaining stable and interpretable survival predictions.

Finally, the limited improvement observed when incorporating sequential structured variables suggests that a large part of the prognostic signal carried by routinely collected structured measurements is already implicitly captured in longitudinal clinical narratives. 
Since these variables must be interpolated over the backward window $[\boldsymbol{\mathrm{L}}-\boldsymbol{\mathrm{w}},\boldsymbol{\mathrm{L}}]$ due to their irregular recording times, this interpolation introduces smoothing assumptions and may propagate measurement noise. 
Consequently, the multimodal configuration provides only limited additional predictive value in this setting.

\section{Conclusion}

We introduced MultiSigBERT, a multimodal survival modeling framework that integrates longitudinal clinical narratives and structured variables within a unified temporal representation. The proposed pipeline combines domain-specific language embeddings concatenated with sequential structured variables, dimensionality reduction, path signature encoding, and sparse Cox regression, yielding a computationally efficient and interpretable approach for survival prediction.

Experiments on a large real-world oncology cohort show that MultiSigBERT achieves strong predictive performance under a landmark design, with a C-index around 0.743 (sd 0.029) and stable calibration. Importantly, the landmark framework ensures a well-defined dynamic prediction task and prevents temporal information leakage by restricting feature construction to observations available prior to the prediction time, therefore providing a robust and reproducible pipeline for longitudinal survival modeling.

The proposed architecture naturally supports multimodal integration. By embedding heterogeneous clinical data into synchronized trajectories prior to signature encoding, the model captures temporal interactions across modalities in a principled geometric representation. In addition, the aggregation strategy enables the use of an arbitrary number of clinical reports per patient while preserving temporal ordering. The entire pipeline remains computationally lightweight and can be trained on standard CPU hardware, offering a favorable trade-off between predictive performance, interpretability, and computational cost.

Some limitations should nevertheless be acknowledged. The structured variables considered in this study remain limited, and their irregular sampling requires interpolation within the landmark window. Future work will investigate additional static covariates, such as tumor type, sex, or age at diagnosis, which can be straightforwardly incorporated into the proposed framework.

Overall, MultiSigBERT provides a robust and scalable framework for multimodal survival analysis from longitudinal clinical data. A key contribution of this work lies in the use of path signatures as a unified representation of heterogeneous temporal modalities. Signature transforms encode temporal dependencies and higher-order interactions into a fixed-dimensional vector while naturally handling irregularly sampled trajectories, making them particularly suitable for real-world clinical data.

Our results also suggest that classical sparse survival models remain highly competitive when combined with expressive temporal representations. The combination of landmarking, signature-based feature extraction, and Cox--LASSO regularization achieved a favorable trade-off between predictive performance, interpretability, computational efficiency, and robustness. Future work will explore richer multimodal representations and external validation on independent oncology cohorts.

\begin{credits}

\subsubsection{\discintname}
The authors declare no potential conflicts of interest.
\end{credits}
%
%
%

\section*{Supplementary}

\begin{figure}[H]
\centering
\includegraphics[width=\textwidth]{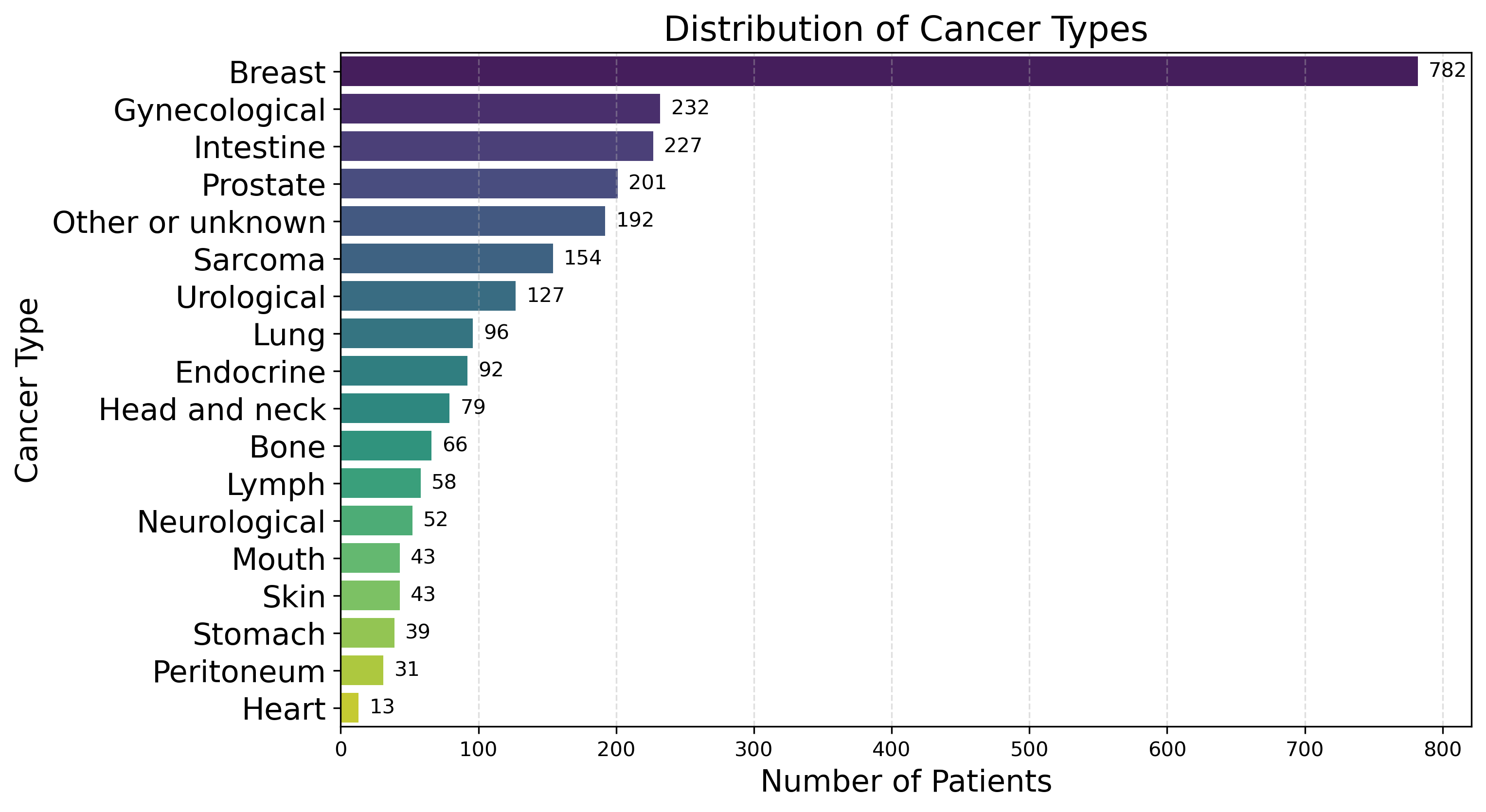}
\caption{
Distribution of cancer types in the study cohort. 
The figure reports the number of patients for each cancer category. 
Breast cancer represents the largest group (782 patients), followed by gynecological (232), intestinal (227), and prostate cancers (201). 
The remaining categories correspond to less frequent tumor types, reflecting the heterogeneous composition of the oncology cohort.
}
\label{fig:cancer_distribution}
\end{figure}

\begin{figure}[H]
\centering
\includegraphics[width=\textwidth]{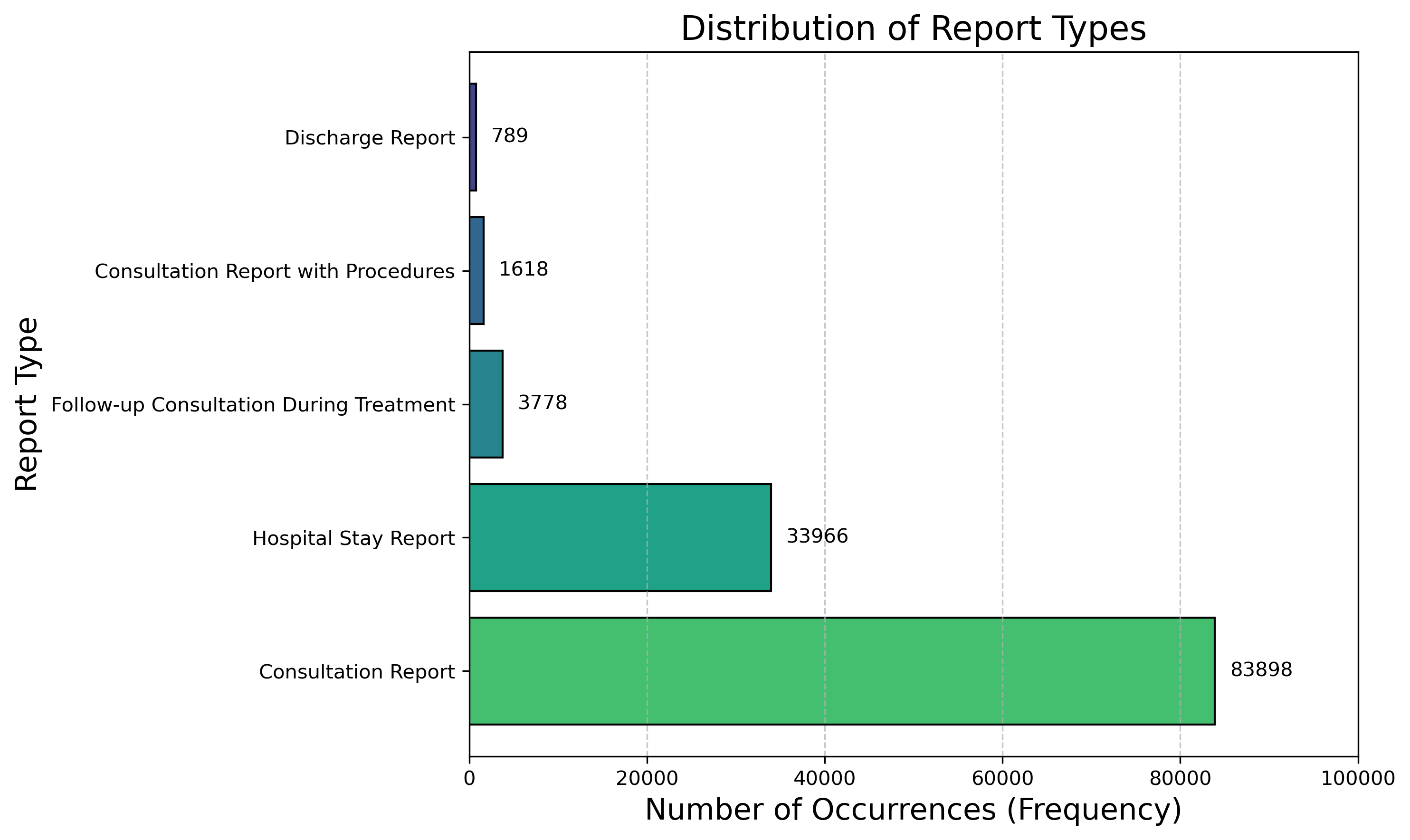}
\caption{
Distribution of report types in the clinical corpus. 
The majority of documents correspond to consultation reports (83,898), followed by hospital stay reports (33,966). 
Additional categories include follow-up consultations during treatment (3,778), consultation reports with procedures (1,618), and discharge reports (789). 
This distribution reflects the predominance of outpatient consultations in routine oncology follow-up, while hospitalizations represent a smaller but clinically important subset of the longitudinal documentation.
}
\label{fig:report_type_distribution}
\end{figure}

\pagebreak
\begin{figure}[H]
\centering
\includegraphics[width=\textwidth]{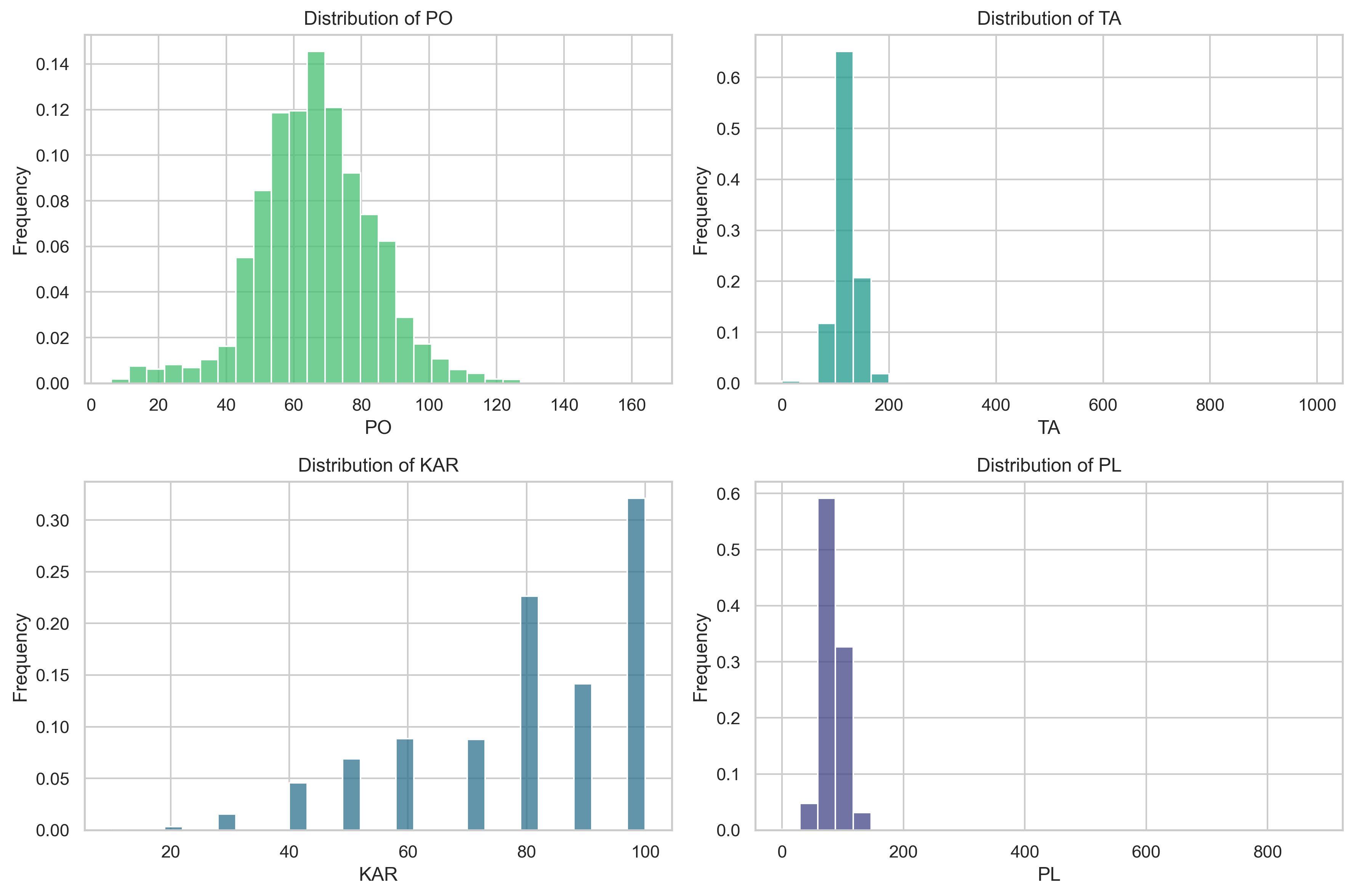}
\caption{
Empirical distributions of the sequential structured clinical variables used in the study: peripheral oxygen saturation (PO), systolic blood pressure (TA), Karnofsky performance index (KAR), and pulse rate (PL). 
Histograms display relative frequencies computed over all available observations in the dataset. 
These variables correspond to longitudinal measurements recorded during patient follow-up and illustrate the heterogeneous sampling patterns and value ranges typical of real-world electronic health records.
}
\label{fig:structured_data_hist}
\end{figure}

\begin{figure}[H]
\centering
\includegraphics[width=\textwidth]{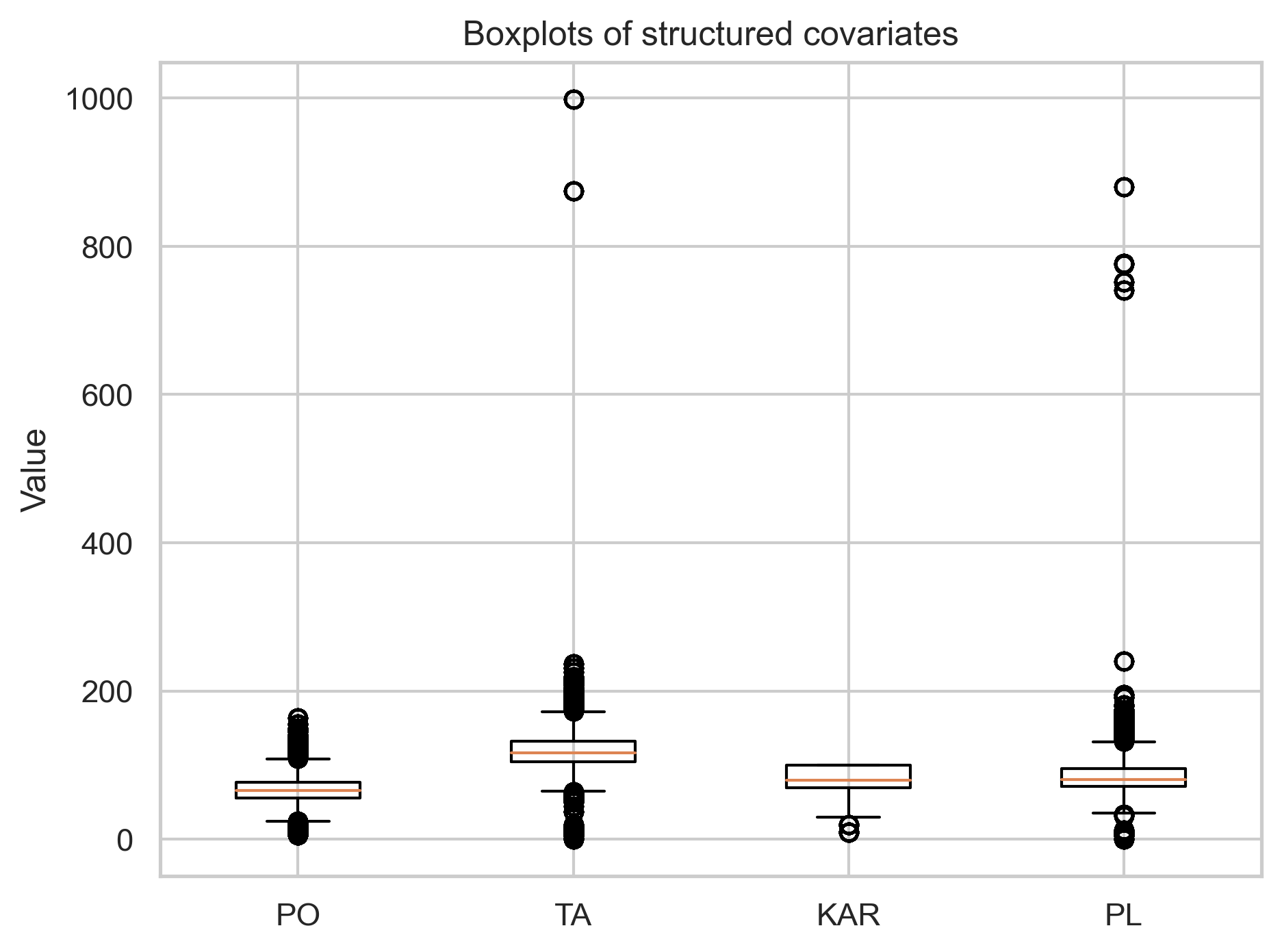}
\caption{
Boxplots of the sequential structured clinical variables used in the study: peripheral oxygen saturation (PO), systolic blood pressure (TA), Karnofsky performance index (KAR), and pulse rate (PL). 
The plots summarize the distribution of values across all observations, showing the median, interquartile range, and potential outliers. 
These measurements are extracted from longitudinal electronic health records and illustrate the variability and occasional extreme values typical of real-world clinical data.
}
\label{fig:structured_data_boxplots}
\end{figure}

\begin{table}[H]
\centering
\begin{tabular}{|l||r|r|r|r|r|r|r|r|r|}
\hline
Variable & Count & NaN Ratio & Mean & Std & Min & Q1 & Median & Q3 & Max \\
\hline
PO (Weight)  & 5,391,366 & 0.316 & 66.38 & 16.97 & 6   & 56  & 66  & 77  & 164 \\
TA (Blood Pressure) & 2,028,745 & 0.743 & 118.88 & 21.44 & 0   & 105 & 117 & 132 & 998 \\
KAR (Karnosky) & 4,453,251 & 0.435 & 80.23 & 19.48 & 10  & 70  & 80  & 100 & 100 \\
PL (Pulse Rate) & 2,019,331 & 0.744 & 83.26 & 18.87 & 0   & 71  & 81  & 95  & 880 \\
\hline
\end{tabular}
\caption{Descriptive statistics of the sequential structured variables. The dataset contains $7{,}885{,}925$ observations corresponding to $2{,}527$ unique patients.}
\label{tab:structured_descriptive_stats}
\end{table}

\pagebreak

\begin{table}[H]
\centering
\begin{adjustbox}{angle=90}
\begin{tabular}{|
>{\raggedright\arraybackslash}m{3.2cm}|
>{\raggedright\arraybackslash}m{4.7cm}|
>{\raggedright\arraybackslash}m{4.1cm}|
m{4.1cm}|
m{1.6cm}|
m{1.6cm}|
m{1.9cm}|}
\hline
\textbf{Model} & \textbf{Field / Dataset} & \textbf{Architecture} & \textbf{Sequential Features} & \textbf{C-index} & \textbf{td-AUC} & \textbf{IBS} \\ \hline
\textbf{SigBERT (\cite{minchella2025sigbert})} & Oncology, narrative reports (Léon Bérard) & OncoBERT + Signature + Cox LASSO & Yes, NLP with path signatures & 0.75 & 0.80 & $\ll$0.25 \\ \hline
\textbf{MSK-CHORD} \cite{Jee2024} & Oncology, Real-world (MSK-CHORD) & Random Survival Forest (RSF) & No, features at fixed time point & $[0.58,0.83]$ & – & – \\ \hline
\textbf{CoxSig} \cite{CoxSig} & Maintenance, synthetic + real (NASA, Califrais) & Cox model + Signature transforms & Yes, time-series encoded with Signature & – & $[0.74,0.87]$ & $[0.09,0.15]$ \\ \hline
\textbf{BERTSurv} \cite{bertsurv} & ICU (MIMIC-III, not oncology) & Transformer (BERT) & Yes, from sequential clinical notes (NLP) & 0.7 & – & – \\ \hline
\textbf{DySurv} \cite{dysurv2024} & ICU (MIMIC-III, eICU) & CVAE + LSTM & Yes, sequential EHR (structured) & $\approx$0.60 & – & included \\ \hline
\textbf{Survival Seq2Seq} \cite{survival_seq2seq} & General (MIMIC-IV + synthetic) & Seq2Seq (GRU-D + Attention) & Yes, hospital time series & – & $[0.84,0.91]$ & – \\ \hline
\textbf{Dynamic-DeepHit} \cite{dynamicdeephit2019} & Cystic Fibrosis (UK Registry) & Deep RNN + Temporal Attention & Yes, repeated biomarker vectors & $[0.94, 0.96]$ & td-AUC & – \\ \hline
\textbf{DeepSurv} \cite{Katzman_2018} & General + oncology (e.g., METABRIC) & DNN with Cox PH loss & No, static baseline covariates & $[0.61, 0.86]$ & – & – \\ \hline
\textbf{Landmark Endpoint} \cite{Devaux2022} & Liver disease (PBC), Aging (PAQUID) & Landmark (Cox, RSF, penalized) & Yes, repeated biomarker measures & – & $[0.73, 0.87]$ & $[0.076, 0.089]$ \\ \hline
\textbf{Penalized Reg. Calib.} \cite{Signorelli_2021} & Neuromuscular (DMD, MARK-MD) & Penalized Cox + Mixed Effects & Yes, blood biomarker sequences & $[0.7, 0.8]$ & $[0.73, 0.87]$ & – \\ \hline
\end{tabular}
\end{adjustbox}
\caption{Overview of representative survival models across domains.}
\label{table:prevmeth}
\end{table}

\end{document}